\documentclass[10pt, a4paper]{article}
\usepackage{lrec-coling2024}
\usepackage{graphicx}
\usepackage{amsmath}
\usepackage{colortbl}

\title{Large Language Models Offer an Alternative to the Traditional Approach of Topic Modelling}

\name{Yida Mu, Chun Dong, Kalina Bontcheva, Xingyi Song} 

\address{Department of Computer Science, The University of Sheffield \\
         \{y.mu, cdong8, k.bontcheva, x.song\}@sheffield.ac.uk\\}
\abstract{
Topic modelling, as a well-established unsupervised technique, has found extensive use in automatically detecting significant topics within a corpus of documents. However, classic topic modelling approaches (e.g., LDA) have certain drawbacks, such as the lack of semantic understanding and the presence of overlapping topics. In this work, we investigate the untapped potential of large language models (LLMs) as an alternative for uncovering the underlying topics within extensive text corpora. To this end, we introduce a framework that prompts LLMs to generate topics from a given set of documents and establish evaluation protocols to assess the clustering efficacy of LLMs. Our findings indicate that LLMs with appropriate prompts can stand out as a viable alternative, capable of generating relevant topic titles and adhering to human guidelines to refine and merge topics. Through in-depth experiments and evaluation, we summarise the advantages and constraints of employing LLMs in topic extraction.
 \\ \newline  }

\begin{document}

\maketitleabstract

\section{Introduction}

Understanding the topics within a collection of documents is crucial for various academic, business, and research disciplines \citep{ramage2009topic,vayansky2020review}. Gaining insights into the primary topics can help in organising, summarising, and drawing meaningful conclusions from vast amounts of textual data.
\footnote{Accepted at LREC-COLING 2024.} 

Classic approaches to topic analysis including 1) \textbf{topic modelling}: an unsupervised approach used to identify themes or topics within a large corpus of text by analysing the patterns of word occurrences \citep{blei2003latent,grootendorst2022bertopic}; and 2) \textbf{close-set topic classification}: model trained on sufficient labelled data with pre-defined close-set topics \citep{wang2012baselines,song2021classification,antypas2022twitter}. However, \textcolor{black}{these} approaches have certain limitations and challenges. \textcolor{black}{Topic classification requires a predefined, closed set of topics and is unable to capture unseen topics.} 
\textcolor{black}{While topic modelling} 
might produce very broad topics while missing out on nuanced or more specific sub-topics that might be of interest (i.e., topic granularity) \citep{abdelrazek2023topic}. Besides, the topics generated by models such as LDA and BERTopic are clusters of words with associated probabilities. Sometimes these clusters might not make intuitive sense to human interpreters, leading to potential misinterpretations \citep{gillings2023interpretation}.

\textcolor{black}{In addition, }these approaches do not perform well on handling unseen documents without a complete re-run of the model, which consequently makes it less efficient for dynamic datasets that are frequently updated (e.g., a dynamic Twitter corpus) \citep{blei2006dynamic,wang2008continuous}.

Addressed those limitations, we proposed an alternative topic modelling approach in this paper -- \textbf{Topic Extraction using Large Generative Language Models}

Generative transformer-based large language models (LLMs) \citep{vaswani2017attention}, such as GPT \citep{brown2020language} and LLaMA \citep{touvron2023llama,touvron2023llama2}, have obtained significant attention for their proficiency in understanding and generating human-like languages. Prompt-based LLMs are transforming conventional natural language processing (NLP) workflows \citep{brown2020language} from model training to evaluation protocols.
For example, existing LLMs (e.g., GPT-4) trained using reinforcement learning with human feedback (RLHF) \citep{ziegler2019fine,ouyang2022training} has shown compatible zero-shot classification performance against supervised methods (e.g., a fully fine-tuned BERT \citep{devlin2019bert}) in various natural language understanding tasks (e.g., detecting customer complaints) in computational social science \citep{ziems2023can,mu2023navigating}.

Due to their plug-and-play convenience, LLMs bring transformative potential to topic modelling. Given that LLMs have the strong capabilities of zero-shot text summarising on par with human annotators \citep{zhang2023benchmarking}, we argue LLMs might leverage their inherent understanding of language nuances to extract (or generate) topics. Their ability to comprehend context, nuances, and even subtle thematic undertones has been demonstrated in various NLP tasks, allowing for a richer and more detailed categorisation of topics \citep{wu2023large,tang2023evaluating}. Besides, the prompt-based model inference pipeline allows users to add manual instructions to guide the model in generating customised outputs \citep{ouyang2022training}. Furthermore, LLMs can seamlessly adapt to evolving language trends and emerging topics (e.g., streaming posts on Twitter), ensuring that topic modelling remains relevant and up-to-date.

\textcolor{black}{Given the lack of prior work, we shed light on the following research aims ranging from model inference and evaluation protocol:}
\begin{itemize}
    \item \textcolor{black}{(i) Investigate the suitability of LLMs as a straightforward, plug-and-play tool for topic extraction without the necessity of complex prompts.}
    \item \textcolor{black}{(ii) Identify and address the limitations and challenges encountered in utilising LLMs for topic extraction.}
    \item \textcolor{black}{(iii) Assess the ability of LLMs to consistently adhere to human-specified guidelines in generating topics with desired granularity.}
    \item \textcolor{black}{(iv) Develop an evaluation protocol to measure the quality of topics generated by LLMs.}
\end{itemize}

To this end, we make the following contributions:
\begin{itemize}
    \item By conducting a series of progressive experiments\footnote{Our source code: \url{https://github.com/GateNLP/LLMs-for-Topic-Modeling}} using different sets of prompting and manual rules, we observed that LLMs with appropriate prompts can be a strong alternative to traditional approaches of topic modelling.
    
    \item We empirically show that LLMs are capable of not just generating topics but also condensing overarching topics from their outputs. The resulting topics, complete with explanations, are easily understood by humans.
    
    \item We introduce evaluation metrics to assess the quality of topics organically produced by LLMs. These metrics are suitable for labelled or unlabelled datasets.
    
    \item Finally, a case study is provided to show the applications of LLMs in real-world scenarios (e.g., analysing topic trends over time). We demonstrate that LLMs can independently perform topic extraction and generate explanations for analysing temporal corpus from a dynamic Twitter dataset (See Figure \ref{fig:case1}).
\end{itemize}

\section{Related Work}
\subsection{Topic Modelling}
\textcolor{black}{
Topic modelling, as a classic unsupervised machine learning approach in computer science, has been broadly employed in various fields such as social science and bio-informatics for processing large-scale of documents \citep{blei2003latent,song2021classification, grootendorst2022bertopic}. A standard output of a topic modelling algorithm is a set of fixed or flexible numbers of topics, where each topic can be typically represented by a list of top words. One can use manual or automatic methods to interpret a topic with corresponding top tokens (e.g., assign a meaningful name for each topic) \citep{lau2010best,allahyari2015automatic}. However, topic interpretation is not always straightforward \citep{aletras2014labelling}. For example, the practice of assigning labels through an eyeballing approach often leads to incomplete or incorrect topic labels \citep{gillings2023interpretation}. Furthermore, topic labelling and interpretation rely heavily on the specialised knowledge of annotators \cite{lee2017human}. Besides, preprocessing (e.g., stemming and lemmatisation) can significantly affect topic modelling performance \citep{chuang2015topiccheck,schofield2016comparing}. Therefore, the use of topic modelling often requires text pre-processing on the input documents and post-processing on the model outputs (e.g., topic labelling) to make the results human-interpretable \citep{vayansky2020review}.
}
\subsection{Close-set Topic Classification}

\textcolor{black}{
On the other hand, closed-set topic classification serves as an alternative to unsupervised topic modelling approaches, which usually depend on models trained on datasets with predefined labels. Topic classification approaches have been widely applied on various domains such as computational social science \citep{wang2012baselines,iman2017longitudinal} and biomedical literature categorisation \citep{lee2006exploring,stepanov2023comparative}. For example, during the COVID pandemic, topic classification approaches were used to analyse the spread of COVID-related misinformation \citep{song2021classification} and public attitudes towards vaccination \citep{poddar2022caves,mu2023examining}.
However, given the nature of the supervised classification task, topic classification typically requires a high cost of human effort in data annotation \citep{antypas2022twitter}. Meanwhile, in the context of labelling social media posts, predefined topics may overlap (such as `News' and `Sports'), leading to disagreements among annotators \citep{antypas2022twitter}.
}

\subsection{LLMs-driven Topic Extraction}
\textcolor{black}{
LLMs have demonstrated their capabilities in text summarisation tasks across various domains, such as news, biomedical, and scientific articles
\citep{wu2023large,shen2023large,tang2023evaluating}. Extractive text summarisation methods can precede LLM-driven topic extraction, i.e., simplifying document complexity and concentrating the topic extraction on the most relevant content. \citep{srivastava2022topic,joshi2023deepsumm}.}

\textcolor{black}{
Meanwhile, LLMs also complement topic modelling approaches, reducing the need for human involvement in the interpretation and evaluation of topics.
\citet{stammbach2023re} explore the use of LLMs for topic evaluation, uncovering that vanilla LLMs (e.g., ChatGPT) can be used as an out-of-the-box approach to automatically assess the coherence of topic word collections. By comparing the human and machine-produced interpretations, \citep{rijcken2023towards} uncover that LLMs ratings highly correlate with human annotations. Besides, topics generated by LLMs are more preferred by general users than the original categories \citep{li2023can}. 
\citet{wang2023large} and \citet{xie2023} point out that LLMs are implicitly topic models which can be used to identify task-related information from demonstrations. 
}

\subsection{Our Work}
In general, previous work has predominantly centred on utilising LLMs as assistants to enhance topic modelling approaches (e.g., automatic evaluation and topic labelling). These studies primarily rely on the output from topic modelling approaches like LDA and BERTopic, instead of topics directly generated by LLMs.
\textcolor{black}{
In this work, we shed light on the potential of using \textbf{LLMs exclusively} for topic extraction and assess topics generated by LLMs from scratch, which is a different task compared to topic modelling and closed-set topic classification.
}

\section{Models and Datasets}
\subsection{LLMs}
In this study, we assess the capability of two widely used LLMs in topic extraction.
\begin{itemize}
    \item \textbf{GPT-3.5 (GPT)}\footnote{\url{https://platform.openai.com/docs/models/gpt-3-5}} represents an advanced iteration of the GPT-3 language model, enhanced with instruction fine-tuning.
Through the OpenAI API, GPT offers plug-and-play capabilities for numerous NLP tasks such as machine translation, common sense reasoning, and question \& answering.
    \item \textbf{LLaMA-2-7B (LLAMA)} \citep{touvron2023llama2} is an enhanced iteration of LLaMA 1 \citep{touvron2023llama}, trained on a corpus that is 40\% larger and with twice the context length. We employ the LLaMA model through the Hugging Face platform\footnote{\url{https://huggingface.co/meta-llama/Llama-2-7b-chat}} \citep{wolf2020transformers}.
\end{itemize}

We chose GPT and LLaMA as they represent two primary modes of LLMs: API-based commercial product and fine-tunable open-source model. Both LLMs have been frequently selected as base models in prior LLM evaluation studies \citep{ziems2023can,mu2023navigating}. Note that there are stronger alternatives such as the GPT-4 and LLaMA-2-70B. However, the chosen models offer more practical implications in terms of financial considerations and computational resources, such as the number of GPUs required and API pricing.

For comparison, we also compare to two widely used baseline models, namely LDA \citep{blei2003latent} and BERTopic \citep{grootendorst2022bertopic}. We utilise LLMs to generate final topic names based on a list of tokens for each topic.

\subsection{Datasets}
To assess the generalisability of LLMs, we examine one open-domain dataset and one domain-specific dataset. We chose these two datasets because they contain texts of varying lengths (i.e., document v.s. sentence levels) and density of vocabulary (i.e., diverse v.s. similar vocabularies). 
Note that topic modelling approaches might struggle with very short texts (e.g., user-generated content on social media) due to the lack of sufficient context to derive meaningful topics. 
Besides, the Twitter dataset also provides temporal information which can be used for analysis topics trends over time (See Case Study in \textbf{\S} \ref{casestudysection}).
\begin{itemize}
    \item \textbf{20 News Group (20NG)} \citep{lang1995newsweeder}, as a classic benchmark\footnote{\url{http://qwone.com/~jason/20Newsgroups/}}, has been widely used in various NLP downstream tasks such as text classification and clustering.

    \item \textbf{CAVS (VAXX)}\footnote{\url{https://github.com/sohampoddar26/caves-data}} \citep{poddar2022caves} is a fine-grained Twitter dataset designed for analysing reasons behind COVID-19 vaccine hesitancy. It contains a collection of tweets labelled under one of ten major vaccine hesitancy categories, such as \textit{`Side-effect' and `Vaccine Ineffective'}.
\end{itemize}
\textbf{Pre-processing} For the Twitter dataset, we perform standard text cleaning rules to filter out all user mentions (i.e., @USER) and hyperlinks. We employ a stratified data split method to sample 20\% of documents from each dataset for the test set, maintaining the same category ratios as in the original dataset.
\section{Experiments}
In this section, we outline our experimental setup, covering both prompt engineering and the strategies we adopted to improve the generation of expected topics. Given the nascent nature of the task, we structure our experiments in a sequence from simpler to more complex prompting settings. This incremental experimental approach aids us in identifying challenges as they arise, guiding us to devise appropriate solutions.

\subsection{Experiment 1: Out-of-box (Basic Prompt)}
We first explore the use of LLMs as an out-of-box approach for topic extraction.
Due to the quadratic complexity of the transformer architecture's attention mechanism with respect to the input sequence length \citep{vaswani2017attention}, LLMs struggle to summarise topics from a large corpus in one prompt.
For instance, even the latest GPT-4\footnote{\url{https://openai.com/research/gpt-4}}, which has expanded its maximum input limit to 32,000 tokens (equivalent to approximately 25,000 words in English), is still incapable of processing most NLP datasets in a single pass.
\subsubsection{Prompting Strategies}
\label{datafeeding}
Consequently, we investigate two prompting strategies: \textbf{(i) feeding text individually and (ii) feeding in batched text} (e.g., 20 documents per batch). 
Note that the former approach incurs slightly higher costs as it necessitates a full prompting message with each iteration.

\begin{figure}[h]
    \centering
    \includegraphics[width=0.48\textwidth]{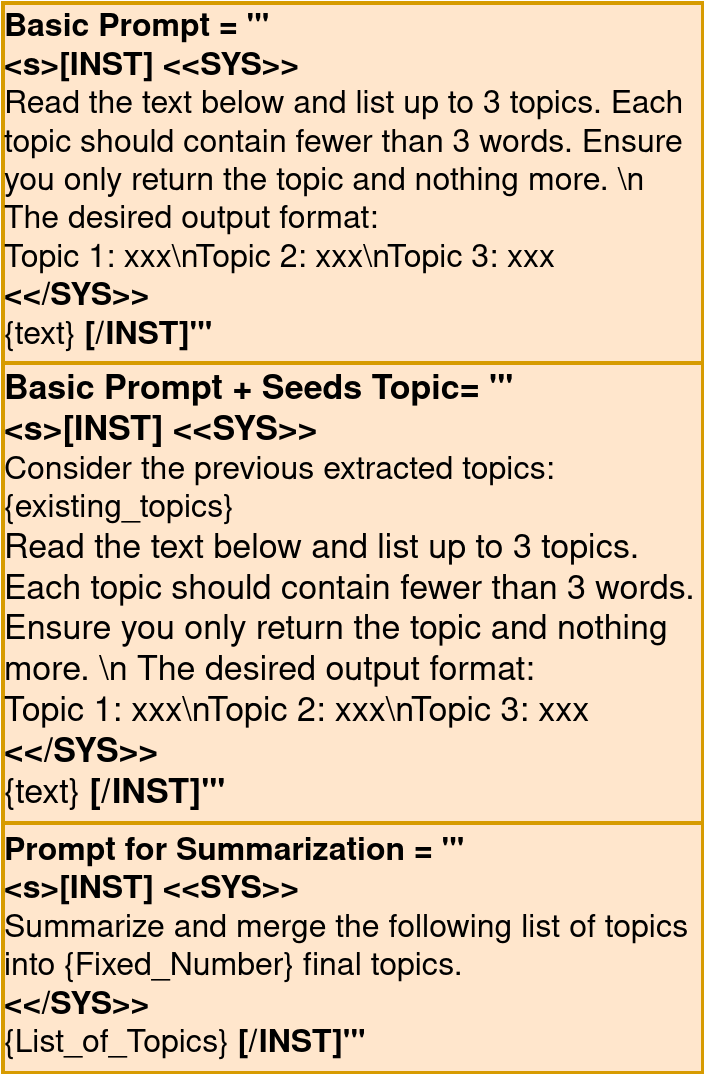} 
    \caption{\small Example prompts for LLaMA. Text enclosed by the special tokens `$\ll$\textit{SYS}$\gg$' is designated as a system prompt.}
    \label{fig:promptexamples} 
\end{figure}

As illustrated in Figure \ref{fig:promptexamples}, our prompt comprises two parts: (i) a system prompt to help the model understand human instructions and the desired output format, and (ii) a user prompt to provide the documents for topic extraction. A structured output format is crucial for subsequent topic statistics and evaluation.

\subsubsection{Results and Discussion}
Given the large number of topics obtained, we count the number of occurrences of each topic and then list the top K topics from the final list. The count of the number of topics represents the proportion of each topic in a given dataset. However, from our initial set of experiments, we observe that both prompting strategies result in similar results. We empirically find that LLMs can reliably handle up to 20 documents per pass depending on the average length.

According to the results of both GPT and LLaMA with `Out-of-box' (See Table \ref{tab:modelresults}, GPT \& LLaMA Expt. 1 Basic Prompt), we observe that LLMs struggle to produce quality topics when given basic instructions without any manual constraints.
Using the VAXX dataset (i.e., fine-grained reasons related to vaccine hesitancy) as an example, we identify the following challenges:

\begin{itemize}
    \item \textbf{Challenge (i)} GPT tends to produce very general topics like \textit{ `Vaccine', `COVID Vaccination' and `Vaccine Hesitancy'}, which are already apparent as the primary themes of the dataset. It suggests that GPT without further manual instructions may not understand the granularity of topics we expected. Note that LLaMA did not return such general with this setting.

    \item \textbf{Challenge (ii)} By manually examining the list of final topics, we also observe a significant overlap in the topics generated by both LLMs, with many topics essentially conveying the same meaning. For example, LLMs might generate topics in various cases and formats, such as \textit{`side-effect', `Side Effect', `serious side effect', `fear of side effects' and `vaccine side effect'.} 

    \item \textbf{Challenge (iii)} Consequently, LLMs generate a large list of topics, which poses a significant challenge in selecting representative topics. From the VAXX dataset, both LLMs return around 2,500 extracted topics, of which 60\% are unique. Given the two existing challenges, simply selecting the top K most frequent topics from the output list of `Experiment Expt. 1 Basic' (See Table \ref{tab:modelresults}) may not yield a truly representative set of topics.
\end{itemize}

\subsubsection{Solutions}
The initial challenges primarily concern obtaining high-quality topics, which are crucial for subsequent topic selection. To this end, we propose the following solutions:
\begin{itemize}
    \item \textbf{Adding Constraints to the Prompt:}
    To avoid the overly broad topics generated by LLMs, we introduce additional constraints in the prompt. For instance, we guide the model not to return broader topics such as `COVID-19' and `COVID-19 Vaccine' in the system prompt. Besides, we also provide task specific information to guide the model to understand the granularity of the given dataset, e.g., by prompting LLMs to return topics related to COVID-19 vaccine hesitancy reasons.

    \item \textbf{Hand-crafted Rules:} We transform similar outputs by adding post-process rules after each iteration. These include using regular expressions to convert all words to lowercase and replace any `Hyphens' with an `Empty Space'. We also apply text lemmatisation rules to normalise all raw outputs (e.g., `vaccine effectiveness' and `vaccine effective').
    
    \item \textbf{Top K Topics} To identify representative topics within the given datasets, we start by employing a simple Top K method, focusing on topics that exhibit the highest frequencies. Further methods for selecting representative topics are discussed in Section \textbf{\S} \ref{set3}.
\end{itemize}
By integrating our proposed solutions into prompts, we achieve improved topic results (see Table \ref{tab:modelresults}, `GPT \& LLaMA Expt. 1 + Manual Instructions') compared to those from `GPT \& LLaMA Expt. 1 + Basic Prompt'.

\begin{table*}[!t]
\resizebox{\textwidth}{!}{%
\begin{tabular}{|l|l|}
\hline
\rowcolor[HTML]{C0C0C0} 
\textbf{\#} &
  \multicolumn{1}{c|}{\cellcolor[HTML]{C0C0C0}\textbf{20 News Group}} \\ \hline
\rowcolor[HTML]{C4F7C3} 
\textit{\textbf{\begin{tabular}[c]{@{}l@{}}Original\\ Categories\end{tabular}}} &
  \textit{\begin{tabular}[c]{@{}l@{}}Comp. {[}graphics, os.ms-windows.misc, sys.ibm.pc.hardware, sys.mac.hardware, windows.x, misc.forsale{]}, \\ Rec. {[}autos, motorcycles, sport.baseball, sport.hockey{]},  Sci. {[}electronics, medical, space, crypt{]}, \\ Soc. {[}religion.christian{]}, Talk. {[}politics.guns, politics.mideast, politics.misc, religion.misc{]}, alt.atheism\end{tabular}} \\ \hline
\rowcolor[HTML]{C4F7C3} 
\textit{\textbf{LDA}} &
  \begin{tabular}[c]{@{}l@{}}Lasting Impressions, People's Perception, Used Cars, New Knowledge, Problem-solving techniques, \\ Armenian Genocide, New System, File Management\end{tabular} \\ \hline
\rowcolor[HTML]{C4F7C3} 
\textit{\textbf{BERTopic}} &
  Game, God, Magnetism, Fire, Depression, Car, Encryption, Server, Technology, Operating System \\ \hline
\textit{\textbf{\begin{tabular}[c]{@{}l@{}}GPT Expt. 1\\ Basic Prompt\end{tabular}}} &
  fbi, faith, baseball, lie, hockey, god, software, cloud, baptism, microsoft \\ \hline
\textit{\textbf{\begin{tabular}[c]{@{}l@{}}GPT Expt. 1\\ + Manual \\ Instructions\end{tabular}}} &
  fbi, faith, baseball, lie, hockey, god, software, microsoft, image conversion, pc \\ \hline
\textit{\textbf{\begin{tabular}[c]{@{}l@{}}GPT Expt. 2\\ + Seeds Topic\end{tabular}}} &
  computer hardware, baseball, religion, hockey, genocide, software, encryption, christianity, faith, price \\ \hline
\textit{\textbf{\begin{tabular}[c]{@{}l@{}}GPT Expt. 3 \\ Summarisation\end{tabular}}} &
  \begin{tabular}[c]{@{}l@{}}Technology \& Computers, Sports, Religion \& Philosophy, Government \& Law, Media \& Entertainment, \\ Health \& Medicine, Vehicles \& Transportation, Society \& Social Issues, History \& Politics, Miscellaneous\end{tabular} \\ \hline
\textit{\textbf{\begin{tabular}[c]{@{}l@{}}LLaMA Expt. 1\\ Basic Prompt\end{tabular}}} &
  technology, future, innovation, price, email, genocide, evidence, sin, books, bible \\ \hline
\textit{\textbf{\begin{tabular}[c]{@{}l@{}}LLaMA Expt. 1\\ + Manual \\ Instructions\end{tabular}}} &
  technology, future, innovation, genocide, bible, god, software, graphics, game, encryption \\ \hline
\textit{\textbf{\begin{tabular}[c]{@{}l@{}}LLaMA Expt. 2\\ + Seeds Topic\end{tabular}}} &
  baseball, hardware, car, software, windows, player, god, government, christianity, hockey \\ \hline
\textit{\textbf{\begin{tabular}[c]{@{}l@{}}LLaMA Expt. 3\\ +Summarisation\end{tabular}}} &
  \begin{tabular}[c]{@{}l@{}}Technology \& Computers, Religion \& Spirituality, Society, Culture, \& Human Rights, Communication \& \\ Media, Sports, Recreation, \& Hobbies, Science \& Research, Economics \& Business, Government, Law, \& \\ Politics, Health \& Wellbeing, Miscellaneous\end{tabular} \\ \hline
\rowcolor[HTML]{C0C0C0} 
\textit{\textbf{\#}} &
  \multicolumn{1}{c|}{\cellcolor[HTML]{C0C0C0}\textbf{Vaccine Hesitancy Reasons}} \\ \hline
\rowcolor[HTML]{C4F7C3} 
\textit{\textbf{\begin{tabular}[c]{@{}l@{}}Original\\ Categories\end{tabular}}} &
  Conspiracy, Country, Ingredients, Mandatory, Pharma, Political, Religious, Rushed, Side effect, Unnecessary \\ \hline
\rowcolor[HTML]{C4F7C3} 
\textit{\textbf{LDA}} &
  \begin{tabular}[c]{@{}l@{}}Vaccine Safety, COVID-19 Vaccination, Vaccine Safety, COVID vaccine, Vaccine effectiveness, COVID-19 \\ Vaccine, Vaccine Allergies, Vaccine Efficacy, COVID-19 Vaccine, COVID-19 Vaccination\end{tabular} \\ \hline
\rowcolor[HTML]{C4F7C3} 
\textit{\textbf{BERTopic}} &
  \begin{tabular}[c]{@{}l@{}}Vaccines, COVID-19, Government spending, COVID, Effectiveness, Untrustworthy, Love, Pandemic, Aging, \\ Influenza, Pandemic, COVID\end{tabular} \\ \hline
\textit{\textbf{\begin{tabular}[c]{@{}l@{}}GPT Expt. 1\\ Basic Prompt\end{tabular}}} &
  \begin{tabular}[c]{@{}l@{}}vaccine hesitancy, vaccine, vaccine effectiveness, vaccine safety, hesitancy, side effects, safety, trust,\\ lack of trust, vaccine efficacy\end{tabular} \\ \hline
\textit{\textbf{\begin{tabular}[c]{@{}l@{}}GPT Expt. 1\\ + Manual \\ Instructions\end{tabular}}} &
  \begin{tabular}[c]{@{}l@{}}vaccine effectiveness, vaccine safety, side effect, trust, vaccine efficacy, misinformation, personal choice, \\ conspiracy theory, fear\end{tabular} \\ \hline
\textit{\textbf{\begin{tabular}[c]{@{}l@{}}GPT Expt. 2\\ + Seeds Topic\end{tabular}}} &
  \begin{tabular}[c]{@{}l@{}}side effect, ineffective, rushed, lack of trust, vaccine safety, vaccine efficacy, death, testing, natural immunity, \\ dangerous, government\end{tabular} \\ \hline
\textbf{\begin{tabular}[c]{@{}l@{}}GPT Expt. 3 \\ Summarisation\end{tabular}} &
  \begin{tabular}[c]{@{}l@{}}Vaccine Efficacy \& Safety, Trust \& Hesitancy, Misinformation \& Beliefs, Government \& Political Influence, \\ Economic \& Financial Concerns, Economic \& Financial Concerns, Vaccine Development \& Availability, \\ Public Perception \& Response, Legal \& Ethical, Comparison \& Alternative, Global \& Societal Impacts\end{tabular} \\ \hline
\textit{\textbf{\begin{tabular}[c]{@{}l@{}}LLaMA Expt. 1\\ Basic Prompt\end{tabular}}} &
  \begin{tabular}[c]{@{}l@{}}safety concerns, lack of trust, misinformation, personal beliefs, side effects, fear of side effects, trust issues, \\ efficacy doubts, personal freedom, effectiveness doubts\end{tabular} \\ \hline
\textit{\textbf{\begin{tabular}[c]{@{}l@{}}LLaMA Expt. 1\\ + Manual \\ Instructions\end{tabular}}} &
  \begin{tabular}[c]{@{}l@{}}safety concerns, lack of trust, misinformation, personal beliefs, side effects, trust issues, efficacy doubts, \\ personal freedom, long term effects, lack of information\end{tabular} \\ \hline
\textit{\textbf{\begin{tabular}[c]{@{}l@{}}LLaMA Expt. 2\\ + Seeds Topic\end{tabular}}} &
  \begin{tabular}[c]{@{}l@{}}side effects, ineffective, safety concerns, lack of trust, efficacy doubts, personal beliefs, effectiveness, \\ misinformation, long term effects, death\end{tabular} \\ \hline
\textit{\textbf{\begin{tabular}[c]{@{}l@{}}LLaMA Expt. 3\\ +Summarisation\end{tabular}}} &
  \begin{tabular}[c]{@{}l@{}}Trust \& Misinformation, Safety \& Side Effects, Efficacy Doubts, Autonomy \& Personal Beliefs, Economic \& \\ Corporate Concerns, Mandatory Vaccination Concerns, Political \& Social Influences, Medical \& Health\\ Concerns,  Access \& Availability, Others\end{tabular} \\ \hline
\end{tabular}%
}
\caption{\small For both LLMs and baseline models, we present the top 10 topics. Additionally, we include the original categories of each dataset for reference.}
\label{tab:modelresults}
\end{table*}

\subsection{Experiment 2: Topics Granularity (GPT \& LLaMA Expt. 2 + Seeds Topic)}
Topic modelling approaches can control the granularity of topics by setting the specific hyper-parameter\footnote{For example, train a LDA via scikit-learn: \url{https://scikit-learn.org/stable/modules/generated/sklearn.decomposition.LatentDirichletAllocation.html}} to fix the number of topics. One can also conduct topic modelling with minimal domain knowledge by adding several anchor words \citep{gallagher2017anchored}.

As shown in Table \ref{tab:modelresults}, the solutions we proposed in experiment 1 can effectively filter out irrelevant topics. However, they fall short in addressing the more complex scenario of similar topics, such as `vaccine side effect', `fear side effect' and `serious side effect'.
Leveraging advanced capabilities in natural language understanding, we test an enhanced prompting setup by using seed topics.
The purpose of providing seed topics is to guide the model towards discerning the granularity of the topics we anticipate. 
This is similar to how a human can understand the potential topics of an unseen set of documents by manually reviewing a few examples to get prior knowledge.
\begin{figure}[t]
    \centering
    \includegraphics[width=0.45\textwidth]
    {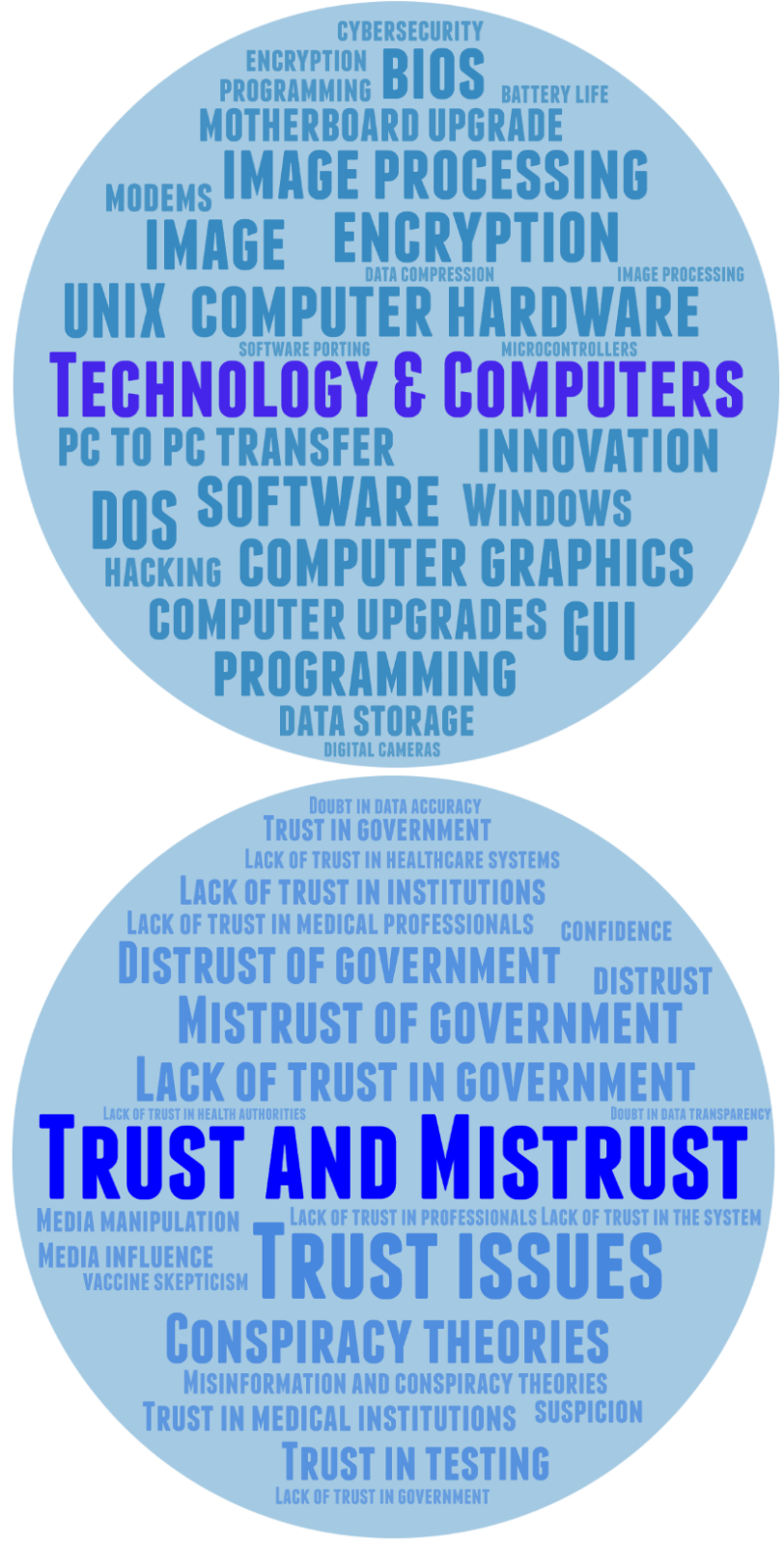} 
    \caption{\small Two examples of final topics summarised by LLMs, namely `Technology \& Computers' (20NG) and `Trust \& Mistrust' (Vaccine dataset).}
    \label{finalexamples} 
\end{figure}
%
\subsubsection{Prompting Strategies}
We select two categories from the original list of labels to serve as seeds topic. They are injected into the prompts to guide the LLMs in understanding the granularity of the topics we anticipate. The example of the prompt demonstrated in the Figure~\ref{fig:promptexamples}, second row ``Basic Prompt + Seeds Topic''.
\subsubsection{Results and Discussion}
In Table \ref{tab:modelresults}, we note that incorporating seed topics (rows titled with Set2 + Seeds Topic) consistently enhances the performance of LLMs in topic extraction across various datasets and LLMs. This indicates that adding seed topics can guide the model in understanding the desired granularity of topics.

\subsection{Experiment 3 Generating Final List (GPT \& LLaMA Expt. 3 + Summarisation)}
\label{set3}
To obtain the final list of topics that can best represent a given set of documents, we consider a further strategy to merge topics into N number of final topics:

\paragraph{Topic Summarisation}
We introduce an additional round of experiments which prompt LLMs to extract the N most appropriate topics from the extracted topic list. While the extracted list is expansive, it is still within the processing capacity of most LLMs, such as the 16k context length of GPT-3.5. For this experiment, we use all raw topics (i.e., the results of GPT \& LLaMA Set 2 + Seed Topics) as input. Through specific prompts, we guide LLMs to produce easily interpretable final N topics with varying granularity. The end result of this process closely mirrors the output format of LDA and BERTopic, where each topic is accompanied by a list of subtopics.

\paragraph{Prompting Strategy}
To guide LLMs in summarising from the final list of topics, we use a prompt that directly asks the model to merge and summarise the given topic list. We also employ a few-shot prompting strategy by manually adding an example to guide the model in generating topics with the desired granularity.  The example of the prompt demonstrated in the Figure~\ref{fig:promptexamples}, third row ``Prompt for Summarisation''.

\paragraph{Results and Discussion}
As indicated in Table \ref{tab:modelresults} (GPT \& LLaMA Expt. 3 + Summarisation), the final set of 10 topics encompasses the majority of the original categories from the source datasets. This highlights the potent proficiency of LLMs in summarising extensive corpora, as demonstrated in prior text summarisation tasks \citep{tang2023evaluating,pu2023Summarization,zhang2023benchmarking}. Moreover, LLMs offer explanations that are easily understandable by humans, detailing the content each topic encompasses. In Figure \ref{finalexamples}, we showcase examples illustrating how LLMs produce interpretable final topics derived from both datasets.

\subsection{Topic Extraction Evaluation}
Previous work has employed evaluation metrics such as perplexity and coherence score \citep{aletras2013evaluating}. 
However, due to the new format of topics generated by LLMs, existing evaluation pipelines are unable to fully handle it. In Table \ref{tab:modelresults}, it is evident that the final Top N list produced by LLMs offers better granularity and interpretability than topics generated using the basic prompt. Nonetheless, having an automated evaluation protocol is crucial for an empirical comparison of model performance. We elucidate our proposed evaluation metrics using outputs from the vaccine dataset:

\begin{figure}[!t]
    \centering
    \includegraphics[width=0.48\textwidth]{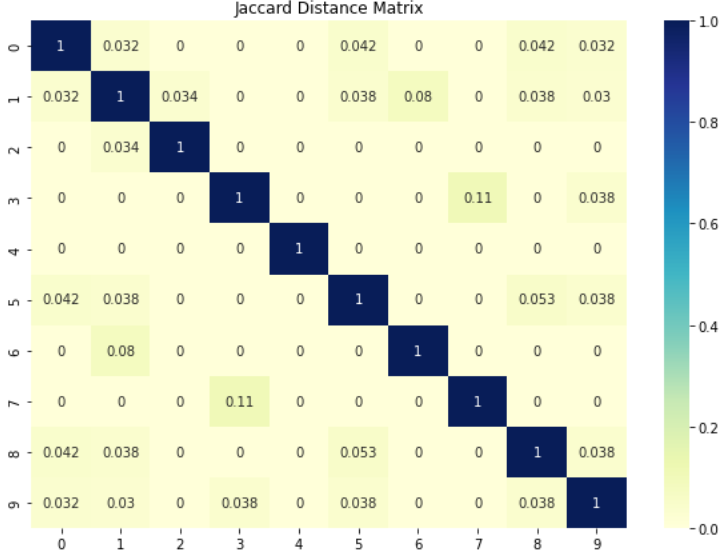} 
    \caption{\small Matrix showing the Jaccard Distance between sub-topics derived from the top 10 general topic in the final list.}
    \label{fig:jaccard} 
\end{figure}

\begin{itemize}
    \item \textbf{(i) Topic Distance over Top N Topics}
     We first use the Jaccard Distance to assess the sub-topics associated with the top N general topics. 
     Given a set of $N$
     general topics, where each topic contains \(10\) sub-topics, we aim to compute the Jaccard distance between each pair of general topics. The Jaccard distance measures dissimilarity between two sets. For two sub-topic lists \(A\) and \(B\), the Jaccard distance is defined as:

    \begin{equation}
        Jaccard\_distance(A, B) = \frac{|A \cap B|}{|A \cup B|}
    \end{equation}

    where:
    \begin{itemize}
        \item \(|A \cap B|\) is the number of elements common to both lists \(A\) and \(B\).
        \item \(|A \cup B|\) is the total number of unique elements across both lists \(A\) and \(B\).
        \item \(Jaccard\_distance(A, B)\) ranges from \(0\) to \(1\), where \(1\) indicates that the lists are identical (i.e., all topics in \(A\) are in \(B\) and vice versa), and \(0\) indicates that the lists share no topics in common.
    \end{itemize}
    Figure \ref{fig:jaccard} shows that the topics in the final list are mostly distinct from one another (topics obtained from the LLaMA Expt. + Summarisation on the VAXX dataset).  
    \item \textbf{(ii) Granularity of Top N Topics}
     We hypothesise that an increased number of topics results in decreased granularity (i.e., higher semantic similarity). To compute the average semantic similarity between each pair of topics from a final top \(N\) topics using the cosine similarity of BERT embeddings \citep{devlin2019bert}, we define the following:

    \begin{itemize}
        \item \( \text{Emb}(T_i) \): BERT embedding (i.e., the `[CLS]' token) of the \(i\)-th topic (768 dimensions).
        \item \( \text{Similar}(T_i, T_j) \): Cosine similarity between the BERT embeddings of the \(i\)-th and \(j\)-th topics.
        \item \( N \): Top N topics.
    \end{itemize}

    Given \(N\) topics, the average semantic similarity between each pair of topics can be computed as follows:

    \begin{itemize}
        \item Compute the BERT embeddings for each topic \( \text{Emb}(T_i) \) for \( i = 1, 2, ..., N \).
        \item Calculate the cosine similarity \( \text{Similar}(T_i, T_j) \) for each unique pair \( (T_i, T_j) \), where \( i \neq j \) and \( i, j = 1, 2, ..., N \).
        \item Compute the average of these similarities obtained above:
            \begin{equation}
           \small
           \text{Ave.} = \frac{2}{N(N-1)} \sum_{i=1}^{N-1} \sum_{j=i+1}^{N} \text{Similar}(T_i, T_j)
    \end{equation}
 
    \end{itemize}

    This equation ensures that each pair is considered only once, as \( \text{similarity}(T_i, T_j) = \text{similarity}(T_j, T_i) \), and there are \( \frac{N(N-1)}{2} \) unique pairs among \(N\) topics. The factor of \(2\) in the numerator adjusts for the fact that we are considering each pair only once in the double summation.

    For our task, we compute the average semantic similarity from the final top N topics, where N takes values of 10, 20, and 30. We notice a positive trend where the average semantic similarity rises with an increase in the number of top N final topics, i.e., Top 10 (0.155), Top 20 (0.197), and Top 30 (0.203). This suggests that LLMs are capable of effectively summarise fine-grained Top N topics when provided with an extensive list of topics.

    \item \textbf{(iii) Recall} Using the seed topics (ST) as a reference, we employ the `Recall' metric to assess how effectively the model can generate pertinent topics (i.e., adhering to human instructions). The recall score is determined by calculating the ratio of correctly identified seed topics to the total number of examples originally labelled as one of the seed topics.
    \begin{equation}
    \small
        \text{Recall} = \frac{\text{No.  Correct Extracted ST Samples}}{\text{No. Seeds Topic Samples}}
    \end{equation}
    \item \textbf{(iv) Precision} Similarly, we compute the precision by determining the ratio of correctly identified seed topics to the total number of examples labelled as a seed topic by LLMs. 
    \begin{equation}
    \small
    \text{Precision} = \frac{\text{No.  Correct Extracted ST Samples}}{\text{No. Samples ST Extracted}}
    \end{equation}
    For the Vaccine dataset, we finally obtain a higher `Recall' (70.0) and a lower `Precision' (49.6) based on the results from the LLaMA Expt. 2 + Seeds Topics.
\end{itemize}

\begin{figure*}[!t]
    \centering
    \includegraphics[width=0.96\textwidth]{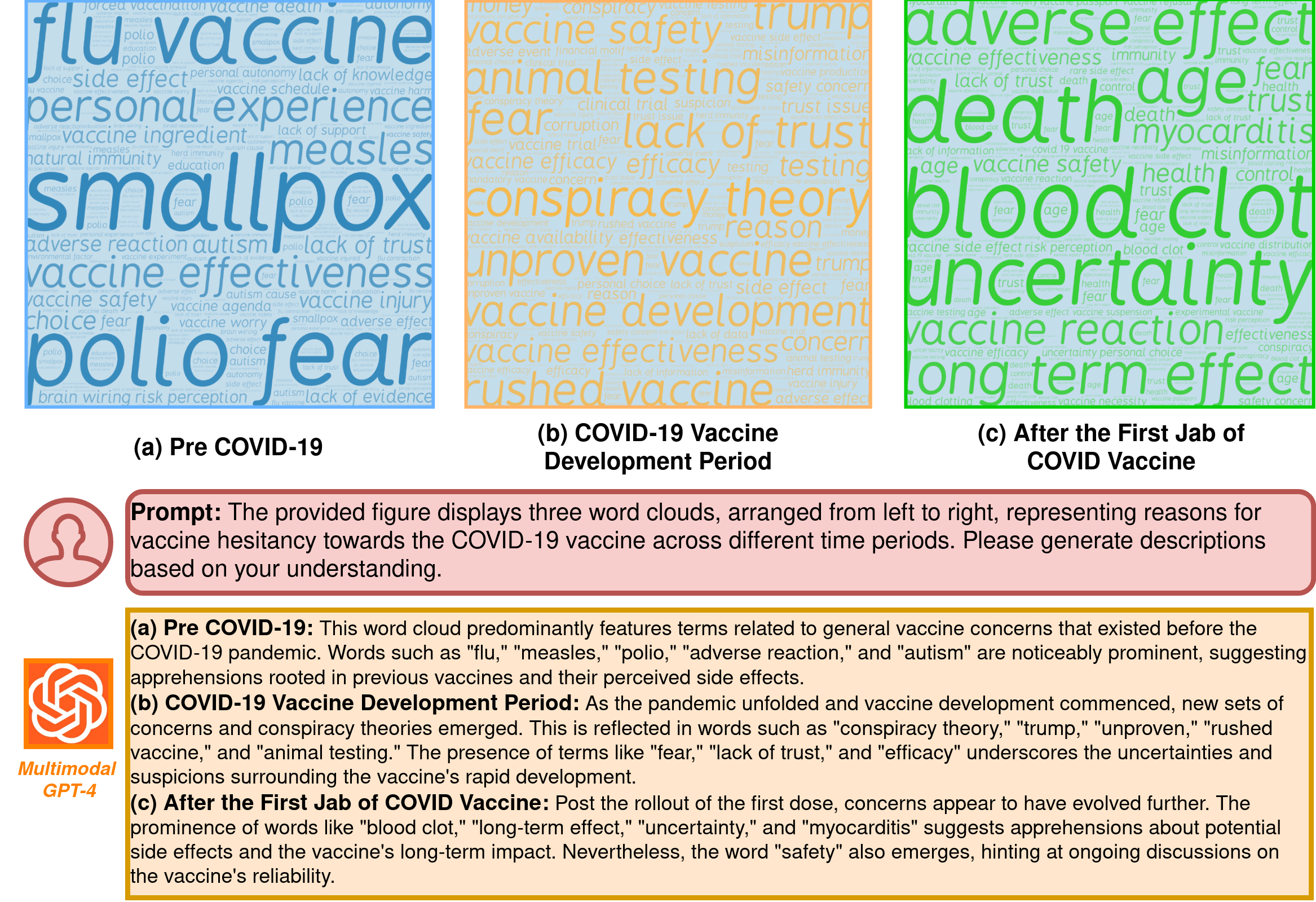} 
    \caption{\small The figure showcases a trio of word clouds (upper half), each capturing the predominant reasons for vaccine hesitancy related to the COVID-19 vaccine over distinct time phases. Additionally, using the cutting-edge multi-modal GPT-4, we obtain descriptions (bottom half) derived from these word clouds.}
    \label{fig:case1} 
\end{figure*}

\section{Case Study: Temporal Analysis of COVID-19 Vaccine Hesitancy}
\label{casestudysection}

Understanding the shifting reasons for hesitancy towards the COVID-19 vaccine is vital, as this knowledge can assist policymakers and biomedical companies in gauging public reactions \citep{poddar2022caves,mu2023vaxxhesitancy}. As time changes, new events, such as emerging reasons for vaccine reluctance, appear that might not have been evident in previous datasets. This indicates that an established LDA or BERTopic model might struggle to process tweets containing these novel topics.
\subsection{Experimental Setup}
We explore the performance of LLMs in addressing unseen topics by processing text in the chronological order. For this purpose, we utilise the Vaccine dataset \citep{poddar2022caves}, as timestamp information is provided in the Twitter metadata. 

Following the timeline of COVID-19 vaccine development\footnote{\url{https://coronavirus.jhu.edu/vaccines}}, we first arrange the Vaccine dataset in chronological order from oldest to latest. We then divide it into three periods:
\textbf{(a)} Pre-COVID-19 (before January 2020),
\textbf{(b)} the COVID-19 vaccine development period (January 2020 to December 2020), and
\textbf{(c)} the period post the first jab of the COVID-19 vaccine when the vaccine became widely adopted globally (after December 2020).\footnote{\url{https://www.rcn.org.uk/magazines/Bulletin/2020/Dec/May-Parsons-nurse-first-vaccine-COVID-19}} 
\subsection{Results and Discussions}
Figure \ref{fig:case1} (top half) showcases the principal topics related to COVID-19 vaccine hesitancy reasons across these three time periods. We also present a description of the figure produced by the cutting-edge multi-model GPT-4 (bottom half).
Our proposed pipeline illustrates that LLMs are capable of automatically executing topic extraction, visualisation (note that LLMs can also generate Python \& R codes for visualising a given set of documents with statistics), and explanation (i.e., based on data visualisation figures). This indicates that both API-based and open source LLMs can serve as a robust substitute for what LDA and BERTopic offer to researchers from various disciplines.

\section{Discussion}
From the standpoint of practical implementation, we summarise the following main takeaways to address our proposed research questions:
\begin{itemize}
    \item 
    Owing to differences in the pre-training corpus and RLHF strategies, various LLMs can exhibit variability in `zero-shot' topic extraction, especially when utilising only basic prompts.
    \item There is no `one-size-fits-all' method of employing LLMs for topic extraction. We recommend conducting preliminary experiments on a small-scale test set. This approach helps early identify potential challenges or issues. 
    \item Upon identifying these limitations, it becomes flexible to establish appropriate constraints and manual guidelines to assist LLMs in topic extraction. Given that LLMs have demonstrated their power in related tasks such as text summarisation and topic labelling, we argue that additional RLHF fine-tuning using a costumed dataset will bolster the LLMs' effectiveness in topic extraction.
    \item By incorporating seed topics, LLMs can generate topics with the desired granularity as specified by users. 
    \item We propose several metrics to assess the quality of topics generated by LLMs from different perspectives, e.g., topic granularity.
\end{itemize}

\section{Conclusion}
In this work, we pioneer the exploration of utilising LLMs for topic extraction. Through empirical testing, we demonstrate that LLMs can serve as a viable and adaptable method for both topic extraction and topic summarisation, offering a fresh perspective in contrast to topic modelling methods. \textcolor{black}{Additionally, LLMs demonstrate their capability to be directly applied to both specific-domain and open-domain datasets for topic extraction.}
This not only underlines the potential of LLMs in understanding hidden topics in large-scale corpora but also opens doors to various innovations (e.g., analysing dynamic datasets) in topic extraction.

\textcolor{black}{
In the future, we plan to concentrate on handling documents that surpass the maximum input length of current LLMs (e.g., LLaMA), for example, by extending the context window of LLMs \citep{chen2023extending,peng2023yarn}. Additionally, we aim to develop new evaluation protocols to directly compare the results from topic modelling approaches and LLM-driven topic extraction, considering the distinct nature of the two tasks.
}
\section*{Ethics Statement}
Our work has been approved by the Research Ethics Committee of our institute and complies with the policies of the Twitter API, the OpenAI API, and Meta LLaMA Terms\&Conditions. 
All datasets are publicly available through the links provided in the original papers. 
All experiments using the OpenAI API cost less than 5 USD, which can be fully covered by the free trial credits.

\section*{Acknowledgements}
This work has been funded by the UK’s
innovation agency (Innovate UK) grant 10039055 (approved under the Horizon Europe Programme as vera.ai\footnote{\url{https://www.veraai.eu/home}}, EU grant agreement 101070093).

\section*{References}
\bibliographystyle{lrec-coling2024-natbib}
\bibliography{lrec-coling2024-example}
\end{document}